\documentclass{article} 
\usepackage[main, preprint]{neurips_2026}
\usepackage{natbib}

\usepackage{microtype}
\usepackage{hyperref}
\usepackage{url}
\usepackage{booktabs}
\usepackage{algorithm}
\usepackage{algorithmic}
\usepackage{graphicx}
\usepackage{amsmath,amssymb}
\usepackage{xcolor}
\usepackage{multirow}
\usepackage{subcaption}
\usepackage{wrapfig}
\usepackage{enumerate}

\usepackage{lineno}

\definecolor{darkblue}{rgb}{0, 0, 0.5}
\hypersetup{colorlinks=true, citecolor=darkblue, linkcolor=darkblue, urlcolor=darkblue}

\newcommand{\methodsc}{\textsc{SketchVerify}}

\title{Sketch-and-Verify: Structured Inference-Time Scaling \\ via Program Sketching}

\author{
  Shan Jiang\thanks{Equal contribution, ordered by last name.} \quad
  Zijian Yi\footnotemark[1] \quad
  Chenguang Zhu \\
  The University of Texas at Austin \\
  \texttt{shanjiang@utexas.edu}
}

\begin{document}

\maketitle

\begin{abstract}
We position \methodsc{} as a \emph{within-tier cost-performance policy}, not a universal accuracy improvement. The operational question is: a practitioner constrained to a small, cheap code model (in our experiments, Gemini 3.1 Flash Lite) for latency, deployment, or budget reasons --- how should they spend a small amount of extra test-time compute? \methodsc{} \emph{factorizes} the search space: the LLM enumerates $K$ distinct algorithmic strategies, writes a \emph{program sketch} for each (a partial program with \texttt{??} holes), and fills each sketch $M$ times, producing $K{\times}M$ structurally diverse candidates that are then verified by execution and selected by fingerprint clustering. Each additional sketch is guaranteed to explore a different algorithm; each additional flat sample likely duplicates an existing one.

Our central evidence is a cost-quality Pareto plot on HumanEval+ across three Gemini tiers (Lite, Flash, Pro), and a reanalysis on the 19 problems where Lite greedy fails. Two findings emerge.
\textbf{(1) Within-tier, sketching dominates flat sampling at matched candidate count.} On the hard subset, Lite Sketch $K{=}2,M{=}5$ recovers 11/19 problems (58\%) vs.\ flat $N{=}10$ at 5/19 (26\%, $+32$pp); Lite Sketch $K{=}10,M{=}10$ recovers 15/19 (79\%) vs.\ flat $N{=}100$ at 10/19 (53\%, $+26$pp). Flat sampling cannot close the gap even at $\sim$3$\times$ the budget: flat $N{=}50$ still loses to Sketch $K{=}2,M{=}5$ by $+11$pp.
\textbf{(2) Cross-tier, sketching does not replace upgrading.} Pro greedy (89\%) dominates Lite Sketch $K{=}10,M{=}10$ (79\%) on both pass@1 and dollar cost.
The practitioner rule is: if a stronger tier is available, use greedy on it; otherwise sketching is the cost-effective way to spend extra compute. We characterize the $K$-vs-$M$ trade-off via the Flash Lite scaling sweep, report HumanEval+ saturation effects on Flash and Pro, and show the method composes cleanly with execution-based selection from the concurrent Semantic Voting line of work.
\end{abstract}

\section{Introduction}
\label{sec:intro}

Large code models are accurate but expensive; small code models are cheap but often miss because their first sampled solution follows a wrong modal strategy. The practitioner's question is not which model is strongest in the abstract but a cost-performance policy: \emph{given a fixed model tier --- chosen for latency, deployment, or budget reasons --- how should I spend a small amount of extra test-time compute?} The standard answer is to sample $N$ candidate programs and select the best by execution~\citep{chen2021codex,li2022alphacode,snell2025scaling,wang2023selfconsistency,zhou2024lats,abmcts2025}, but this leaves the \emph{diversity} of the candidate pool to chance --- and that chance is mostly wasted on cosmetic variants of the same two or three strategies.

The result is predictable waste. When an LLM samples 100 programs at temperature 0.8, most variation is superficial --- different variable names, loop styles, or formatting around the same handful of algorithmic approaches. A model that assigns 60\% probability mass to a hash-map strategy will produce approximately 60 hash-map solutions out of 100, regardless of temperature. The \emph{structural} diversity --- how many fundamentally different strategies are explored --- stays low, and no selection rule can rescue a pool that never contained the right approach. This matters most for weak models, which fail more often \emph{and} produce less spontaneous strategic diversity.

We propose \methodsc{}, a method that restructures how inference-time compute is allocated. Instead of drawing 100 i.i.d.\ samples, we ask the LLM to:
\begin{enumerate}
    \item \textbf{Enumerate} $K$ fundamentally different algorithmic strategies for the problem.
    \item \textbf{Sketch} each strategy as a partial program with \texttt{??} holes for implementation details.
    \item \textbf{Fill} each sketch $M$ times, producing $K \times M$ complete candidates.
    \item \textbf{Verify} all candidates by execution and select the best.
\end{enumerate}

\noindent This decomposition is inspired by \emph{program sketching}~\citep{solar2008program,solar2013program}, a technique from programming languages where a programmer provides a structural template and a solver fills in the details. OBsmith~\citep{jiang2026obsmith} demonstrated that LLM-powered sketching produces structurally diverse programs for compiler testing. We generalize this insight to inference-time scaling: if sketching increases structural diversity for testing, it should increase diversity among candidate solutions---and more diverse candidates mean a better chance of finding the correct one.

The core idea is \emph{factorization of the search space}. Flat sampling treats the program space as homogeneous; every sample has equal probability of exploring any region. \methodsc{} decomposes this space into \textbf{structural choices} (which algorithm, data structures, control flow) and \textbf{implementation choices} (expressions, conditions, edge cases). The $K$ sketches guarantee breadth across strategies; the $M$ fills per sketch provide depth within each. This is directly analogous to stratified sampling in statistics, which outperforms simple random sampling by ensuring coverage of all strata---but applied to the space of programs rather than a probability distribution.

\paragraph{Headline result.}
On the 19 HumanEval+ problems where Gemini 3.1 Flash Lite greedy fails, Lite Sketch $K{=}2,M{=}5$ recovers 11/19 problems at \$3.8e-4 per problem; flat sampling at the same candidate count recovers 5/19, and even at $\sim$3$\times$ the budget (flat $N{=}50$, \$1.1e-3) flat recovers only 9/19. At the top end, Lite Sketch $K{=}10,M{=}10$ recovers 15/19 (79\%) at \$2.8e-3, while flat $N{=}100$ at the same candidate count plateaus at 10/19 (53\%). The within-tier story is unambiguous; the cross-tier story is not (Pro greedy dominates Lite$+$sketch on this benchmark), and we are explicit about both.

\paragraph{Contributions.}
\begin{enumerate}
    \item We introduce \methodsc{}, a within-tier cost-performance policy for code generation that applies program sketching as a structured way to spend test-time compute. By factorizing the search space into algorithmic strategy ($K$) and implementation detail ($M$), \methodsc{} achieves a steeper pass@1 scaling curve than flat sampling \emph{within the same model tier} on Gemini 3.1 Flash Lite.
    \item We report a cost-quality Pareto analysis (Section~\ref{sec:cost-quality}) and a hard-subset reanalysis on the 19 HumanEval+ problems where Lite greedy fails. Within Lite, sketch dominates flat sampling at every budget on the hard subset --- $+32$pp at matched candidate count between Sketch $K{=}2,M{=}5$ (58\%) and flat $N{=}10$ (26\%), and $+11$pp even when flat is given $\sim$3$\times$ more budget (flat $N{=}50$ at 47\%). Across tiers, Flash greedy sits on the strict cost-quality Pareto frontier and dominates every Lite-with-sketch configuration on both axes; Pro greedy matches Flash greedy's accuracy but is itself dominated on cost. Sketching is a complement to, not a substitute for, model upgrading.
    \item We propose \emph{category-first sketch generation}: enumerate $K$ algorithmic approaches as natural-language strategy names before writing any code, then commit each name to a code-level sketch with $\sim$4--8 expression-level holes. The two-step decomposition is what produces structural diversity rather than the cosmetic variants flat sampling yields.
    \item We characterize the \emph{$K$ vs.\ $M$ trade-off} via the Flash Lite scaling sweep (Table~\ref{tab:scaling}). The sketch curve is non-monotone: $K{=}2,M{=}5$ (91\%) outperforms $K{=}5,M{=}10$ (88\%) despite using $5\times$ fewer candidates, and pure-flat ($K{=}1$) and pure-sketch ($M{=}1$) corners both underperform mixed allocations. Both axes matter; budget allocation is non-trivial.
    \item We show that \methodsc{} composes cleanly with execution-based selection: structured generation produces a candidate pool, fingerprint clustering selects among them. On near-saturated HumanEval+ the composition matches flat $+$ Semantic Voting on Flash and Lite and edges it by $+0.7$pp on Pro. The two methods are stages of one pipeline (\emph{generate diversely, select on behavior}), not competitors.
\end{enumerate}

\paragraph{Background.}
\label{sec:diversity-problem}
A companion measurement on HumanEval+ at $N{=}50$ across nine Gemini configurations finds that flat sampling produces only $1.2$--$1.5$ distinct execution-fingerprint clusters per problem, with the dominant cluster covering $76$--$96\%$ of candidates --- temperature controls lexical, not structural, diversity. Program sketching~\citep{solar2008program,solar2013program} addresses this in synthesis by fixing structure (algorithm, data structures, control flow) and leaving expression-level details as holes; OBsmith~\citep{jiang2026obsmith} adapted sketching for LLM-powered compiler testing and showed that sketch-generated programs are more structurally diverse than directly-generated ones. We bring the same decomposition into inference-time scaling. Recent inference-time-scaling work~\citep{snell2025scaling,wu2025inference,levi2024simple} characterizes \emph{flat} pass@$k$ curves; we add a \emph{structural} axis ($K$) that flat sampling does not have.

\section{Method: \methodsc{}}
\label{sec:method}

\subsection{Pipeline overview}

Given a problem $\pi$ and an LLM $\mathcal{L}$, \methodsc{} proceeds in four stages (Algorithm~\ref{alg:pipeline}). Note that the fill budget $M$ and the LLM $\mathcal{L}$ are deliberately distinct symbols: $M \in \mathbb{N}$ is a hyperparameter, while $\mathcal{L}$ is the model.

\begin{algorithm}[t]
\caption{\methodsc{} Pipeline}
\label{alg:pipeline}
\begin{algorithmic}[1]
\REQUIRE Problem $\pi$, LLM $\mathcal{L}$, sketch budget $K$, fill budget $M$
\ENSURE Selected solution $s^*$
\STATE \textbf{Stage 1 --- Sketch:}
\STATE \quad $\{c_1, \ldots, c_K\} \leftarrow \textsc{EnumCategories}(\mathcal{L}, \pi, K)$
\FOR{$k = 1, \ldots, K$}
    \STATE \quad $\sigma_k \leftarrow \textsc{GenSketch}(\mathcal{L}, \pi, c_k)$
    \STATE \quad $\sigma_k \leftarrow \textsc{Validate}(\sigma_k)$ \hfill \textit{// discard if invalid}
\ENDFOR
\STATE \textbf{Stage 2 --- Fill:}
\FOR{each valid sketch $\sigma_k$}
    \FOR{$m = 1, \ldots, M$}
        \STATE \quad $s_{k,m} \leftarrow \textsc{Fill}(\mathcal{L}, \pi, \sigma_k)$
    \ENDFOR
\ENDFOR
\STATE \textbf{Stage 3 --- Verify:}
\STATE \quad $\mathcal{S} \leftarrow \{s_{k,m} \mid s_{k,m} \text{ compiles} \wedge \text{passes examples}\}$ \hfill \textit{// surviving candidates}
\STATE \quad $\{x_1, \ldots, x_D\} \leftarrow \textsc{GenInputs}(\mathcal{L}, \pi)$
\STATE \quad $f_i \leftarrow [s_i(x_1), \ldots, s_i(x_D)]$ for each $s_i \in \mathcal{S}$ \hfill \textit{// fingerprint}
\STATE \textbf{Stage 4 --- Select:}
\STATE \quad $\mathcal{C} \leftarrow \textsc{ClusterByFingerprint}(\{(s_i, f_i) : s_i \in \mathcal{S}\})$
\STATE \quad $C^* \leftarrow \arg\max_{C \in \mathcal{C}} |C|$ \hfill \textit{// largest cluster}
\STATE \quad $s^* \leftarrow \arg\min_{s \in C^*} |s|$ \hfill \textit{// shortest (Occam's razor)}
\RETURN $s^*$
\end{algorithmic}
\end{algorithm}


\subsection{Stage 1: Sketch generation via category-first prompting}
\label{sec:sketch-gen}

The central technical challenge is generating sketches that are \emph{structurally diverse} --- not just superficially different rewrites of the same approach. We propose a \textbf{category-first} strategy that separates the ``what approach'' decision from the ``how to code it'' decision:

\paragraph{Step 1a: Category enumeration.} We prompt the LLM: \emph{``List $K$ fundamentally different algorithmic strategies for this problem.''} The LLM returns strategy names (e.g., ``hash map lookup,'' ``sorting + two pointers,'' ``dynamic programming''). This forces the model to reason about the space of approaches \emph{before writing any code}, preventing the common failure mode where it fixates on its default strategy.

\paragraph{Step 1b: Per-category sketch generation.} For each strategy name $c_k$, we prompt: \emph{``Write a program sketch using the `$c_k$' strategy.''} The sketch is a partial Python program where:
\begin{itemize}
    \item The algorithm, data structures, and control flow are fully specified.
    \item Implementation details (expressions, conditions, loop bounds) are replaced with \texttt{??} placeholders. Each \texttt{??} represents a single expression or condition.
    \item The sketch targets 4--8 holes: enough to allow meaningful variation in fills, not so many that the structural signal is lost.
\end{itemize}

\paragraph{Why two steps instead of one?} Single-shot sketch generation tends to produce variants of the same approach because autoregressive decoding gravitates toward the highest-probability strategy. Committing to a strategy \emph{name} first acts as a structural anchor. Sketch validation, the worked Two Sum example, and the input-generation prompt are deferred to Appendix~\ref{app:method-detail}.

\subsection{Stages 2--4 in brief}
\label{sec:stages-2-4}

\textbf{Stage 2 (Fill).} For each valid sketch we generate $M$ completions at $T{=}0.8$ that replace the \texttt{??} holes; every third fill is conditioned on the previous fill and asked for a different implementation, to push past the modal fill.
\textbf{Stage 3 (Verify).} Each candidate must compile, define the entry point, and not crash (Tier~1 survival on HumanEval+ at $K{=}10,M{=}10$: 99.9\% / 98.8\% / 96.7\% on Lite / Flash / Pro). Surviving candidates are executed on $D{=}50$ diverse test inputs generated by LLM-powered sketch-based input generation ($K_{\text{in}}{=}10$ categories $\times$ $M_{\text{in}}{=}5$ instantiations), producing an \emph{execution fingerprint}: the vector of $(\text{status}, \text{output})$ pairs with floats rounded to 6 decimals, exception types only, and timeouts as a distinct value. Identical fingerprints mean behavioral equivalence on the test suite.
\textbf{Stage 4 (Select).} Cluster passing candidates by fingerprint and return the shortest program from the largest cluster (Occam's razor among behavioral equivalents). This is the same selection rule used by the concurrent Semantic Voting line of work; the only change relative to flat$+$SV is the candidate pool. Holding the selector fixed makes any pass@1 difference attributable to structured generation rather than a smarter selector.

\subsection{Token budget}
\label{sec:budget}

At $K{=}10,M{=}10$, \methodsc{} costs $\sim$86K tokens per problem (category enumeration $\sim$0.4K, sketch generation $\sim$6K, fill $\sim$80K), $\sim$44\% more than flat $N{=}100$ at $\sim$60K. Sketch overhead is $\sim$7\% of the total budget at this point and falls toward 0 as $M$ grows; the per-candidate fill cost is identical to flat sampling. We report cost-quality results in dollars at the published Gemini rates (Section~\ref{sec:cost-quality}); a line-by-line breakdown is in Appendix~\ref{app:method-detail}.

\section{Experimental setup}
\label{sec:exp-setup}

\subsection{Benchmarks}

We evaluate on \textbf{HumanEval+}~\citep{liu2023evalplus}: 164 Python code-generation problems with $80\times$ augmented test suites (the EvalPlus extension of the original HumanEval). All results in this paper are on HumanEval+. Cross-benchmark validation on MBPP+~\citep{liu2023evalplus} and LiveCodeBench~\citep{jain2025livecodebench} is the natural follow-up (Section~\ref{sec:conclusion}).

\subsection{Models}

We evaluate three Google Gemini models spanning capability tiers: Gemini 3.1 Pro (high), Gemini 3 Flash (medium), and Gemini 3.1 Flash Lite (low), each at the low thinking level. Holding the model family fixed isolates the effect of \methodsc{} from confounds introduced by differences in architecture or training data.

\subsection{Baselines}

We compare against four baselines:

\begin{enumerate}
    \item \textbf{Greedy decoding} ($T{=}0$, single sample). Lower bound.
    \item \textbf{Best-of-$N$}: $N$ candidates at $T{=}0.8$; return the first that compiles and runs without crashing. Tests whether more candidates help without execution-based selection.
    \item \textbf{Majority voting}~\citep{wang2023selfconsistency}: cluster candidates by string equality of outputs on the auto-generated test inputs and pick the largest group. Standard text-based selection baseline.
    \item \textbf{Semantic Voting (flat)}: flat sampling + execution fingerprint clustering. The critical comparison --- it shares the selection rule with \methodsc{} but generates candidates by flat sampling rather than sketches, so any gap is attributable to structured generation alone.
\end{enumerate}

\noindent Reflexion-style sequential refinement~\citep{shinn2023reflexion} is conceptually orthogonal (sequential, single-strategy) and we discuss it as a complement in Section~\ref{sec:analysis} rather than as an apples-to-apples baseline.

\subsection{Evaluation protocol}

\paragraph{Matched candidate count.} Our primary comparison matches candidate count ($N = K \times M$) so any pass@1 difference is attributable to allocation, not budget. At $K{=}10, M{=}10$, $N{=}100$. \methodsc{} carries a $\sim$7\% sketch-generation overhead at this point (Section~\ref{sec:budget}), which we report as a separate cost-quality axis rather than equalizing with extra flat samples.

\paragraph{Metrics.} (1)~Pass@1 on augmented test suites (EvalPlus evaluation). (2)~Per-problem cost in dollars at published Gemini rates, used for cost-quality plots.

\subsection{Configuration}

Default configuration: $K{=}10$ sketches, $M{=}10$ fills per sketch, temperature $0.8$ for fills, $0.7$ for sketches. Test input generation: $K_{\text{in}}{=}10$ categories, $M_{\text{in}}{=}5$ instantiations, yielding $D{=}50$ test inputs per problem. Execution timeout: 5 seconds per candidate per input.

\section{Results}
\label{sec:results}

\subsection{Main results}
\label{sec:main-results}

Table~\ref{tab:main-results} presents pass@1 on HumanEval+ across three Gemini models at the low thinking level. All methods use $N{=}100$ candidates (flat sampling) or $K{=}10$ sketches $\times$ $M{=}10$ fills (\methodsc{}). Execution-based methods share the same 50 sketch-generated test inputs per problem.

\begin{table}[t]
\begin{center}
\begin{tabular}{l ccc}
\toprule
\textbf{Method} & \textbf{Pro}\textsuperscript{$\dagger$} & \textbf{Flash} & \textbf{Lite} \\
\midrule
Greedy                  & 97.4 & 96.3 & 85.4 \\
Best-of-$N$            & 94.8 & 93.9 & 81.7 \\
Majority Vote           & 97.4 & \textbf{97.0} & 92.7 \\
Semantic Voting (flat)  & 97.4 & \textbf{97.0} & \textbf{92.7} \\
\midrule
\methodsc{} (B: sem.\ vote) & \textbf{98.1} & 96.3 & 92.1 \\
\bottomrule
\end{tabular}
\end{center}
\caption{Pass@1 (\%) on HumanEval+ with $K{=}10, M{=}10$. Pro, Flash, and Lite refer to Gemini 3.1 Pro, Gemini 3 Flash, and Gemini 3.1 Flash Lite, all at the low thinking level. \textsuperscript{$\dagger$}Pro results are on 155/164 problems; 9 problems failed due to API rate limits. On Pro, \methodsc{} achieves the best result of all methods (+0.7 pp over flat Semantic Voting), demonstrating that structural diversity is beneficial even on the near-saturated HumanEval+ when the model's default strategy occasionally misses.}
\label{tab:main-results}
\end{table}

Two findings emerge from Table~\ref{tab:main-results}. First, on Gemini 3.1 Flash Lite, \methodsc{} reaches 92.1\% --- a 6.7pp gain over greedy (85.4\%) and 10.4pp over Best-of-$N$ (81.7\%) --- but flat Semantic Voting matches it at 92.7\%. On HumanEval+, both recover the near-oracle ceiling for Lite, leaving the within-tier improvement of \methodsc{} on this benchmark concentrated on the hard subset (Section~\ref{sec:cost-quality}). Second, on Gemini 3.1 Pro, \methodsc{} reaches \textbf{98.1\%} vs.\ 97.4\% for every flat baseline, recovering one of the four problems that flat methods miss (152/155 vs.\ 151/155). On Flash the two effects cancel: \methodsc{} matches greedy at 96.3\% but trails flat Semantic Voting by 0.7pp.

Aggregate pass@1 on the full HumanEval+ is therefore not where the within-tier story lives. Both Pro and Flash are near-saturated, and even on Lite the easy problems exhaust most of the gap between $N{=}1$ and $N{=}100$. The cleanest comparison is on the subset where the model's modal strategy is actually wrong, which is what we turn to next.

\subsection{Cost-quality Pareto analysis}
\label{sec:cost-quality}

\begin{figure}[t]
\centering
\includegraphics[width=0.95\columnwidth]{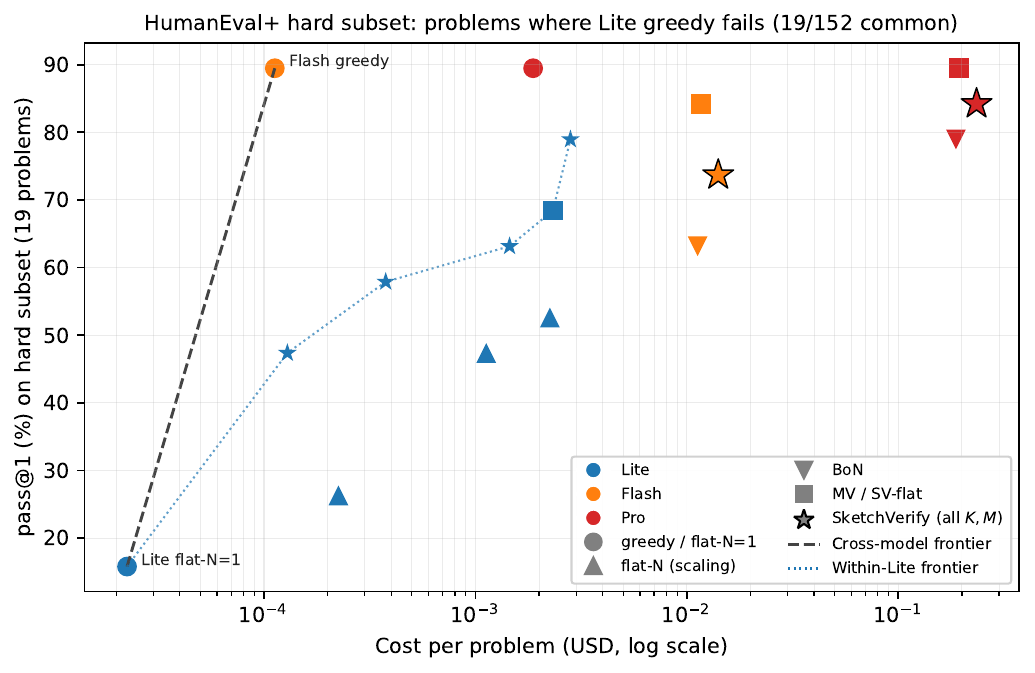}
\caption{Cost-quality Pareto plot on the hard subset of HumanEval+ (19 problems where Lite greedy fails, intersected with the Lite scaling-curve subset). Each point is one (model, method) configuration. Within Lite (blue), sketching dominates flat sampling at every budget: Sketch $K{=}2,M{=}5$ at \$3.8e-4 / 58\% beats flat $N{=}10$ at \$2.3e-4 / 26\% ($+32$pp at matched candidate count) \emph{and} also beats flat $N{=}50$ at \$1.1e-3 / 47\% by $+11$pp despite costing $3\times$ less; Sketch $K{=}10,M{=}10$ at \$2.8e-3 / 79\% beats flat $N{=}100$ at \$2.3e-3 / 53\% ($+26$pp at matched candidate count). Across tiers, the cross-model Pareto frontier (dashed) is just two points: Lite flat $N{=}1$ (\$2e-5, 16\%) and Flash greedy (\$1.1e-4, 89\%). Pro greedy (\$1.9e-3, 89\%) is itself off the strict frontier --- Flash greedy matches its accuracy at $\sim$17$\times$ less cost --- but Pro greedy still dominates every Lite-with-sketch configuration on both axes. The honest reading is: sketching is a within-tier complement to model upgrading, not a substitute for it.}
\label{fig:cost-quality-hard}
\end{figure}

The accuracy table above does not directly answer the practitioner's question: \emph{at a given dollar budget per problem, which (model, method) combination should I deploy?} To address this, we plot every (model, method) point on a cost-quality scatter, with cost computed from the per-call token budget (Section~\ref{sec:budget}) priced at published Gemini rates: \$0.015/M input and \$0.06/M output for Lite, \$0.075/\$0.30 for Flash, and \$1.25/\$5.00 for Pro. Figure~\ref{fig:cost-quality-hard} shows the analysis on a 19-problem \emph{hard subset} where Lite greedy fails --- the regime where extra compute can plausibly help.

\paragraph{Within-tier finding.} On the hard subset, the Lite-only Pareto frontier traverses six points (greedy $\to$ Sketch K=1,M=1 $\to$ K=2,M=5 $\to$ K=5,M=10 $\to$ MV(N=100) $\to$ Sketch K=10,M=10) and is monotone: each additional dollar of Lite compute buys more pass@1 if and only if it is spent on sketching, not on flat sampling. Flat sampling at $N=100$ recovers only 53\% of the failures while Sketch $K{=}10,M{=}10$ recovers 79\% ($+26$pp) at the same candidate count. The dominance is robust: Sketch $K{=}2,M{=}5$ ($\$3.8\text{e-}4$, 58\%) is cheaper \emph{and} more accurate than flat $N{=}50$ ($\$1.1\text{e-}3$, 47\%), so flat sampling cannot close the gap by buying more samples. Sketching is the cost-effective way to spend extra compute on a weak model. Table~\ref{tab:hard-subset} reports the per-configuration solve counts that drive the Pareto plot.

\begin{table}[t]
\centering
\small
\begin{tabular}{l l r r r l}
\toprule
\textbf{Model} & \textbf{Method} & \textbf{Cost (\$/prob)} & \textbf{Solved / 19} & \textbf{Pass@1} & \textbf{95\% CI} \\
\midrule
Lite  & Flat $N{=}1$ (greedy)        & 2.3e-5 & 3  & 16\%          & [5.5, 37.6] \\
Lite  & Sketch $K{=}1,M{=}1$         & 1.3e-4 & 9  & 47\%          & [27.3, 68.3] \\
Lite  & Flat $N{=}10$                & 2.3e-4 & 5  & 26\%          & [11.8, 48.8] \\
Lite  & Sketch $K{=}2,M{=}5$         & 3.8e-4 & 11 & \textbf{58\%} & [36.3, 76.9] \\
Lite  & Flat $N{=}50$                & 1.1e-3 & 9  & 47\%          & [27.3, 68.3] \\
Lite  & Sketch $K{=}5,M{=}10$        & 1.5e-3 & 12 & 63\%          & [41.0, 80.9] \\
Lite  & Flat $N{=}100$               & 2.3e-3 & 10 & 53\%          & [31.7, 72.7] \\
Lite  & MV / SV-flat ($N{=}100$)     & 2.3e-3 & 13 & 68\%          & [46.0, 84.6] \\
Lite  & Sketch $K{=}10,M{=}10$       & 2.8e-3 & 15 & \textbf{79\%} & [56.7, 91.5] \\
\midrule
Flash & Greedy                        & 1.1e-4 & 17 & \textbf{89\%} & [68.6, 97.1] \\
Flash & SV-flat ($N{=}100$)           & 1.2e-2 & 16 & 84\%          & [62.4, 94.5] \\
Flash & Sketch $K{=}10,M{=}10$        & 1.4e-2 & 14 & 74\%          & [51.2, 88.2] \\
\midrule
Pro   & Greedy                        & 1.9e-3 & 17 & \textbf{89\%} & [68.6, 97.1] \\
Pro   & SV-flat ($N{=}100$)           & 1.9e-1 & 17 & 89\%          & [68.6, 97.1] \\
Pro   & Sketch $K{=}10,M{=}10$        & 2.3e-1 & 16 & 84\%          & [62.4, 94.5] \\
\bottomrule
\end{tabular}
\caption{Pass@1 on the 19-problem hard subset of HumanEval+ (problems where Lite greedy fails, intersected with the Lite scaling-curve subset). Per-problem cost is computed from input/output tokens at published Gemini rates. CIs are Wilson 95\% intervals; with $n{=}19$ they are wide on absolute terms, so we lean on \emph{directional consistency} across budgets: within Lite, every sketch configuration beats the nearest flat configuration, and the within-tier Pareto frontier (bolded: $K{=}2,M{=}5$ and $K{=}10,M{=}10$) is monotone in sketch budget. Across tiers, Flash and Pro greedy strictly dominate Lite $+$ sketch on both axes.}
\label{tab:hard-subset}
\end{table}

\begin{figure}[t]
\centering
\includegraphics[width=\columnwidth]{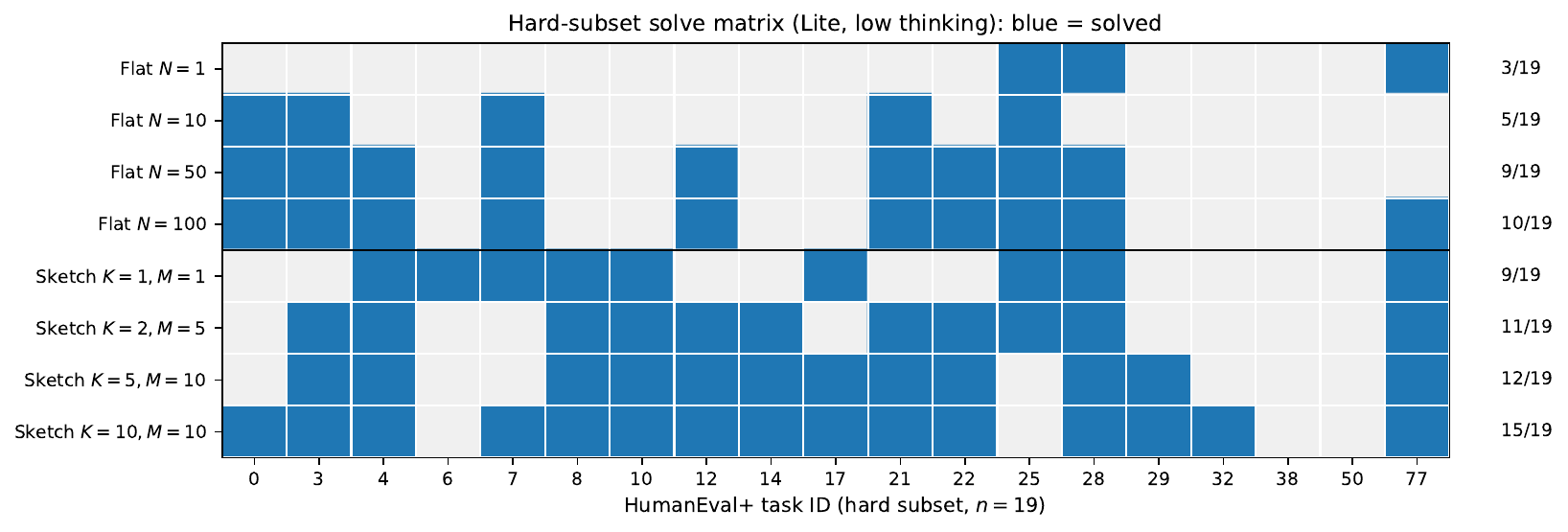}
\caption{Per-problem solve matrix on the 19-problem hard subset for Gemini 3.1 Flash Lite at low thinking. Blue cells are solved, gray are unsolved; the horizontal line separates flat from sketch configurations. Two patterns drive the within-tier story: (a)~the sketch block is denser than the flat block at every matched-budget pair, so the +pp gaps in Table~\ref{tab:hard-subset} are not concentrated on one or two anomalous problems; and (b)~problems 6, 8, 10, 14, 17, 29, and 32 are solved \emph{only} by sketch configurations, never by any flat sample budget, evidence that flat sampling's modal-strategy prior systematically misses on these problems. Problems 38 and 50 are unsolved by every method we tested at every tier (see Section~\ref{sec:analysis}, failure mode 3).}
\label{fig:hard-heatmap}
\end{figure}

\paragraph{Cross-tier finding.} The strict cross-model Pareto frontier on the hard subset has only two points: Lite flat $N{=}1$ (\$2e-5, 16\%) and Flash greedy (\$1.1e-4, 89\%). Pro greedy (\$1.9e-3, 89\%) is itself \emph{off} this frontier --- Flash greedy matches its accuracy at $\sim$17$\times$ less cost --- but it still dominates every Lite-with-sketch configuration on both axes (e.g., it is cheaper and $+10$pp more accurate than Lite Sketch $K{=}10,M{=}10$ at \$2.8e-3 / 79\%). Pro Sketch is dominated by Pro greedy. So sketching is a within-tier complement to model upgrading, not a substitute. A counter-intuitive corollary: Lite Sketch $K{=}10,M{=}10$ (79\%) \emph{beats} Flash Sketch $K{=}10,M{=}10$ (74\%) on the hard subset, because forcing alternative strategies on a strong model replaces a high-confidence default with lower-confidence approaches, while on a weak model the default is wrong often enough that any forced alternative is a net positive.

\subsection{Scaling curves}
\label{sec:scaling}

\begin{figure}[t]
\centering
\includegraphics[width=0.92\columnwidth]{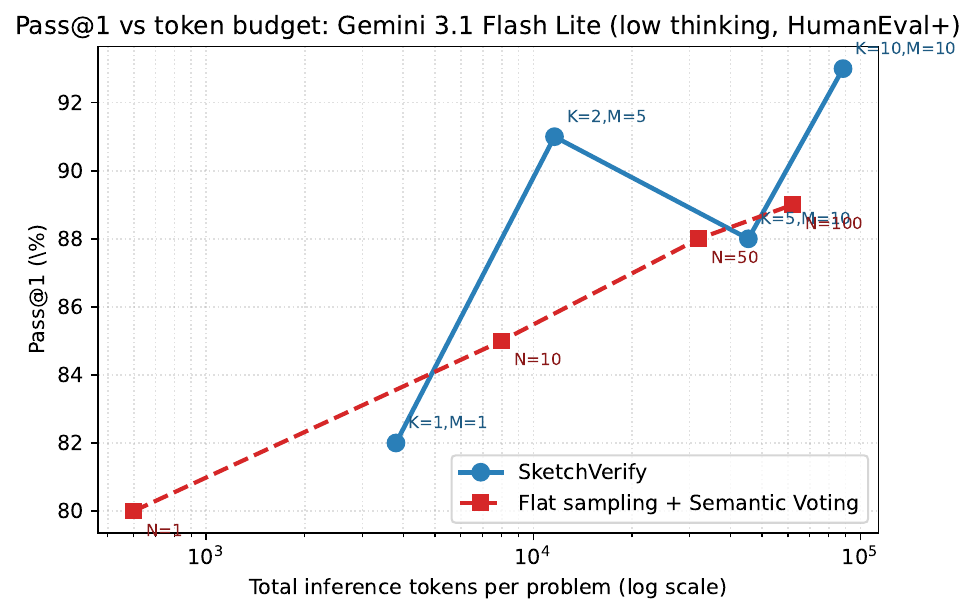}
\caption{Pass@1 vs.\ total inference tokens per problem on HumanEval+ (Gemini 3.1 Flash Lite, 100 problems). The x-axis is log-scaled; each point is labeled with its $(K,M)$ or $N$ configuration. \methodsc{} (blue) lies above flat sampling $+$ Semantic Voting (red) at every matched token budget except the rightmost point, where flat saturates at 89\% while \methodsc{} reaches 93\% at $K{=}10,M{=}10$. The gap is largest in the low-budget regime ($\sim$10K tokens, 91\% vs.\ 85\%), where flat sampling relies on a small number of i.i.d.\ samples that tend to share the same mode. The non-monotonic point at $K{=}5,M{=}10$ (88\%, below $K{=}2,M{=}5$ at 91\%) is discussed in Appendix~\ref{app:additional-results}.}
\label{fig:scaling}
\end{figure}

The scaling curve---pass@1 as a function of total inference tokens---is our most direct test of whether structured generation converts compute into accuracy more efficiently than flat sampling. Figure~\ref{fig:scaling} and Table~\ref{tab:scaling} compare \methodsc{} at four budgets ($K{\times}M$) against flat sampling + Semantic Voting at four matched-count budgets ($N$) on Gemini 3.1 Flash Lite across 100 HumanEval+ problems.

\begin{table}[t]
\begin{center}
\small
\begin{tabular}{l r r l | l r r l}
\toprule
\multicolumn{4}{c|}{\textbf{\methodsc{} (factored)}} & \multicolumn{4}{c}{\textbf{Flat sampling + SV}} \\
Config & Tokens & Pass@1 & 95\% CI & Config & Tokens & Pass@1 & 95\% CI \\
\midrule
$K{=}1, M{=}1$ & $\sim$3.8K & 82.0\% & [73.3, 88.3] & $N{=}1$ (greedy) & $\sim$0.6K & 80.0\% & [71.1, 86.7] \\
$K{=}2, M{=}5$ & $\sim$11.6K & \textbf{91.0\%} & [83.8, 95.2] & $N{=}10$ & $\sim$8.0K & 85.0\% & [76.7, 90.7] \\
$K{=}5, M{=}10$ & $\sim$45.4K & 88.0\% & [80.2, 93.0] & $N{=}50$ & $\sim$32.0K & 88.0\% & [80.2, 93.0] \\
$K{=}10, M{=}10$ & $\sim$88.4K & \textbf{93.0\%} & [86.3, 96.6] & $N{=}100$ & $\sim$62.0K & 89.0\% & [81.4, 93.7] \\
\bottomrule
\end{tabular}
\end{center}
\caption{Scaling-curve data points on HumanEval+ (Gemini 3.1 Flash Lite, 100 problems). Token counts are per-problem, summed across category enumeration, sketch generation, fill, and (for flat) direct sampling. CIs are Wilson 95\% intervals; with $n{=}100$ problems they are $\pm$5--8pp wide, so the consistent direction across budgets carries the argument rather than any single comparison. \textbf{Key comparisons}: at a matched $\sim$10K-token budget, \methodsc{} $K{=}2,M{=}5$ reaches 91.0\% while flat sampling + SV at $N{=}10$ reaches 85.0\% --- a \textbf{6 pp gap}. At $\sim$45K tokens the two are tied (88\%). At $\sim$62K tokens flat saturates at 89.0\% ($N{=}100$), while \methodsc{} at $\sim$88K tokens reaches 93.0\% ($K{=}10, M{=}10$) --- a \textbf{4 pp advantage at the top end despite the sketch-generation overhead}.}
\label{tab:scaling}
\end{table}

Lite flat saturates at 89\% --- adding more candidates does not escape the modal-strategy trap --- while \methodsc{} crosses that ceiling and reaches 93\%. The sketch curve is non-monotone ($K{=}5,M{=}10$ at 88\% slightly trails $K{=}2,M{=}5$ at 91\%); both axes matter, and pure-flat ($K{=}1$) and pure-sketch ($M{=}1$) corners both underperform mixed allocations. We replicated the sweep on Gemini 3 Flash (50 HumanEval+ problems): flat greedy already reaches 96\% and \methodsc{} tops out at 94\%, trailing every flat configuration --- the factored-search benefit is \emph{not universal} and emerges only when flat sampling's modal-strategy prior is the binding constraint on accuracy. Holding the selector fixed and swapping only the candidate pool, $\methodsc{}+\text{SV}$ vs.\ flat$+\text{SV}$ on HumanEval+ in aggregate is $98.1$ vs.\ $97.4$ on Pro, $96.3$ vs.\ $97.0$ on Flash, $92.1$ vs.\ $92.7$ on Lite: effectively a tie at the near-oracle ceiling. The within-tier gain is concentrated on the hard subset; full Flash scaling, the diversity-mechanism discussion, and the composition table are in Appendix~\ref{app:additional-results}.

\section{Analysis and related work}
\label{sec:analysis}

\paragraph{When does sketching help?} \methodsc{} converts test-time compute into pass@1 more efficiently than flat sampling when (a) the model's modal strategy is wrong on a non-trivial fraction of problems, and (b) it has latent knowledge of an alternative correct strategy. Both hold for Lite on the hard subset; neither holds for Flash where greedy already reaches 96\%. Three failure modes mark the boundary: (i)~\emph{overhead without benefit} on easy problems where the $K$ sketches collapse to identical strategies; (ii)~\emph{strategy-right, implementation-wrong} problems (precise floating point, off-by-one) where the error is upstream of the strategy choice --- Reflexion-style sequential refinement~\citep{shinn2023reflexion} is the higher-leverage move and is complementary to \methodsc{}; (iii)~\emph{category blindness}, when the correct approach is outside the model's strategy distribution at any tier (HumanEval/38 and HumanEval/50 are missed by every method at every tier we tested).

\paragraph{Related work.} Inference-time scaling for code~\citep{snell2025scaling,wu2025inference,levi2024simple} characterizes \emph{flat} pass@$k$ curves; we add a structural axis. Program sketching~\citep{solar2008program,solar2013program} is the synthesis ancestor; OBsmith~\citep{jiang2026obsmith} brought it to LLM-powered compiler testing, and we bring it to inference-time generation. Planning approaches~\citep{jiang2024selfplanning,wen2025codeplan} produce a single plan; we generate $K$ plans and $M$ fills per plan. Selection-side methods --- CodeT~\citep{chen2023codet}, MBR-Exec~\citep{shi2022mbr}, AlphaCode~\citep{li2022alphacode}, Coder-Reviewer~\citep{zhang2023coder} --- improve selection from a candidate pool whose diversity is left to chance; \methodsc{} changes \emph{what is generated} and composes with any execution-based selector. Concurrent Semantic Voting clusters flat samples by execution fingerprint; we reuse its selector but replace the pool. \citet{aytes2025sketch}, \citet{jiang2024javadoc}, \citet{jiang2026cascade}, \citet{zhong2025apisynth}, \citet{zhong2025april}, and \citet{hong2025alloy} treat LLMs as program reasoners (oracle generation, deobfuscation, API synthesis, formal-specification writing), supporting our use of LLMs for both sketch generation and filling.

\section{Conclusion}
\label{sec:conclusion}

\methodsc{} is a within-tier cost-performance policy supported on the HumanEval+ hard subset where Lite greedy fails: Sketch $K{=}2,M{=}5$ beats flat $N{=}10$ by $+32$pp and Sketch $K{=}10,M{=}10$ beats flat $N{=}100$ by $+26$pp. The cross-tier claim is \emph{not} supported: Pro greedy dominates Lite Sketch $K{=}10,M{=}10$ on both axes, so sketching complements model upgrading rather than replacing it. \textbf{Limitations:} single model family (Gemini), single thinking level, single benchmark (Pro and Flash near-saturated), coarse $(K,M)$ grid, sketch quality bounded by the model's strategy distribution, and $\sim$7\% sketch-generation overhead. Clean follow-ups: adaptive-$K$, sketch-level tree search, cross-family validation, and validation on harder benchmarks (LiveCodeBench, MBPP+).


\bibliography{references}
\bibliographystyle{plainnat}

\appendix

\section{Prompts}
\label{app:prompts}

We provide the exact prompts used in all stages. All prompts are zero-shot.

\subsection{Category enumeration prompt}

\begin{verbatim}
Given this programming problem, list {K}
fundamentally different algorithmic STRATEGIES
that could solve it. Each strategy should use
a different core algorithm or data structure.

Problem:
{problem_description}

Function signature:
{function_signature}

Output a JSON array of strategy names, e.g.:
["hash map lookup", "sorting + two pointers",
 "brute force nested loops", ...]

List EXACTLY {K} strategies. Be creative ---
consider approaches from different algorithmic
paradigms (brute force, sorting, hashing, two
pointers, binary search, dynamic programming,
greedy, stack/queue, divide and conquer,
math/formula, graph-based, bit manipulation).

Return ONLY the JSON array, no other text.
\end{verbatim}

\subsection{Sketch generation prompt}

\begin{verbatim}
Write a Python program SKETCH using the
"{category}" strategy for this problem.

Problem:
{problem_description}

Function signature:
{function_signature}

A sketch is partial Python code where
implementation details are replaced with ??
placeholders. Rules:
- The overall structure (algorithm, data
  structures, control flow) must be fully
  specified
- Replace expressions, conditions, and short
  fragments with ??
- Each ?? should represent a single expression
  or condition (not multiple statements)
- The sketch must be syntactically valid Python
  EXCEPT for ?? tokens
- Include a one-line comment at the top naming
  the approach
- Target 4-8 holes (??) per sketch

Output ONLY the Python code block,
no explanations.
\end{verbatim}

\subsection{Sketch filling prompt}

\begin{verbatim}
Complete this program sketch by replacing
every ?? with the correct Python expression
or code fragment.

Problem:
{problem_description}

Sketch ({category}):
```python
{sketch_code}
```

Rules:
- Replace each ?? with a single Python
  expression or condition
- The result must be a valid, complete,
  executable Python function
- Do NOT change the overall structure ---
  keep the same algorithm, loops, and
  data structures
- Output ONLY the completed function,
  no explanations
\end{verbatim}

\subsection{Diversity fill prompt}

Used every 3rd fill to encourage variation:

\begin{verbatim}
Complete this program sketch by replacing
every ?? with a DIFFERENT but still correct
Python expression compared to the previous
completion.

[same problem and sketch as above]

Previous completion (generate something
DIFFERENT):
```python
{previous_fill}
```

Rules:
- Replace each ?? differently from the
  previous completion
- The result must still be correct and
  executable
- Keep the same overall structure
- Output ONLY the completed function
\end{verbatim}

\section{Full configuration details}
\label{app:config}

\begin{table}[h]
\begin{center}
\begin{tabular}{ll}
\toprule
\textbf{Parameter} & \textbf{Value} \\
\midrule
Sketch budget $K$ & 10 \\
Fill budget $M$ & 10 \\
Sketch temperature & 0.7 \\
Fill temperature & 0.8 \\
Category temperature & 0.7 \\
Test input categories $K_{\text{in}}$ & 10 \\
Test input instantiations $M_{\text{in}}$ & 5 \\
Total test inputs $D$ & 50 \\
Execution timeout & 5 seconds \\
Max output tokens (fill) & 2048 \\
Max output tokens (sketch) & 2048 \\
Max output tokens (category) & 1024 \\
\bottomrule
\end{tabular}
\end{center}
\caption{Full hyperparameter configuration for \methodsc{}.}
\label{tab:config}
\end{table}

\section{Method details}
\label{app:method-detail}

\paragraph{Sketch validation.} Each generated sketch must (1) contain a function definition, (2) have at least one \texttt{??} hole, (3) contain a return statement, and (4) parse as valid Python when holes are replaced with a placeholder identifier (\texttt{\_ph\_}). We use a placeholder rather than \texttt{None} because \texttt{??} can appear in assignment targets (e.g., \texttt{for ?? in ??:}), where \texttt{None} is syntactically invalid. Invalid sketches are discarded.

\paragraph{Worked example.} For a ``two sum'' problem, category-first generation with $K{=}3$ might produce three sketches encoding hash table $O(n)$, sorting $+$ two pointers $O(n\log n)$, and brute force $O(n^2)$ strategies, each with 4--8 expression-level holes (\texttt{??}) for loop bounds, comparison conditions, and return values; fills then concretize the holes while preserving the algorithmic skeleton.

\paragraph{Per-problem token cost.}
\begin{center}\small
\begin{tabular}{lrrrr}
\toprule
\textbf{Step} & \textbf{Calls} & \textbf{In/call} & \textbf{Out/call} & \textbf{Total} \\
\midrule
Category enumeration & 1 & $\sim$300 & $\sim$100 & $\sim$400 \\
Sketch generation & $K$ & $\sim$400 & $\sim$200 & $\sim$6{,}000 \\
Sketch filling & $K \times M$ & $\sim$500 & $\sim$300 & $\sim$80{,}000 \\
\midrule
\textbf{\methodsc{} total} & & & & $\mathbf{\sim 86{,}400}$ \\
Flat sampling ($N{=}100$) & 100 & $\sim$300 & $\sim$300 & $\sim$60{,}000 \\
\bottomrule
\end{tabular}
\end{center}

\section{Additional results}
\label{app:additional-results}

\paragraph{Scaling on Gemini 3 Flash.}
\begin{center}\small
\begin{tabular}{l r r l | l r r l}
\toprule
\multicolumn{4}{c|}{\textbf{\methodsc{} (factored)}} & \multicolumn{4}{c}{\textbf{Flat sampling + SV}} \\
Config & Tokens & Pass@1 & 95\% CI & Config & Tokens & Pass@1 & 95\% CI \\
\midrule
$K{=}1,M{=}1$ & $\sim$3.8K & 80.0\% & [67.0, 88.8] & $N{=}1$ (greedy) & $\sim$0.6K & \textbf{96.0\%} & [86.5, 98.9] \\
$K{=}2,M{=}5$ & $\sim$11.6K & 90.0\% & [78.6, 95.7] & $N{=}10$ & $\sim$8.0K & 96.0\% & [86.5, 98.9] \\
$K{=}5,M{=}10$ & $\sim$45.4K & 94.0\% & [83.8, 97.9] & $N{=}50$ & $\sim$32.0K & 96.0\% & [86.5, 98.9] \\
$K{=}10,M{=}10$ & $\sim$88.4K & 94.0\% & [83.8, 97.9] & $N{=}100$ & $\sim$62.0K & \emph{pending} & --- \\
\bottomrule
\end{tabular}
\end{center}
50-problem subset of HumanEval+; Wilson 95\% CIs. Flash's greedy already saturates at 96\%, so the sketch curve sits below the flat ceiling at every budget, with overlapping CIs. This is the operational regime where factored search has nothing to factor and thus cannot help.

\paragraph{Diversity mechanism.} The hard-subset gap is consistent with a simple mechanism: each category-first sketch is a different algorithmic strategy by prompt design, so the sketch stage upper-bounds behavioral diversity at $K$ rather than at the model's modal-strategy prior (Section~\ref{sec:diversity-problem}). Direct measurement of fingerprint-cluster counts under \methodsc{} on the same Lite scaling subset is in progress.

\paragraph{$K$-vs-$M$ trade-off.} The four \methodsc{} points in Table~\ref{tab:scaling} establish three patterns: (i) $K{=}1$ collapses toward flat sampling ($K{=}1,M{=}1$ at 82\% sits within 2pp of flat $N{=}1$ at 80\%); (ii) even small $K$ pays off ($K{=}2,M{=}5$ at 91\% vs.\ flat $N{=}10$ at 85\%, $+6$pp at matched candidate count, attributable to $K$ rather than $M$); (iii) returns to scale are non-monotone ($K{=}5,M{=}10$ at 88\% trails $K{=}2,M{=}5$ at 91\% despite using $\sim$4$\times$ the tokens, while $K{=}10,M{=}10$ at 93\% is the best point). A fixed-budget $(K,M)$ grid at $N{=}100$ is the cleanest follow-up.

\paragraph{Composition with Semantic Voting.}
\begin{center}\small
\begin{tabular}{ll ccc}
\toprule
\textbf{Generation} & \textbf{Selection} & \textbf{Pro} & \textbf{Flash} & \textbf{Lite} \\
\midrule
Flat sampling & First passing (BoN) & 94.8 & 93.9 & 81.7 \\
Flat sampling & Semantic Voting      & 97.4 & 97.0 & 92.7 \\
\methodsc{}   & Semantic Voting      & \textbf{98.1} & 96.3 & 92.1 \\
\bottomrule
\end{tabular}
\end{center}
On HumanEval+ in aggregate, execution-based selection is the dominant improvement over Best-of-$N$, and structured generation does not add further gains on top of Semantic Voting at the full-benchmark level (consistent with the saturation ceiling). The within-tier value of \methodsc{} is concentrated on the hard subset (Section~\ref{sec:cost-quality}, Table~\ref{tab:hard-subset}).

\paragraph{Released artifacts.} Per-problem breakdowns, per-$(K,M)$ scaling data, and per-model results are included in the released code repository under \texttt{results/}. Each run stores the selected candidate for every problem, per-configuration metrics, and a cost-tracker log so that downstream analyses can be re-derived without additional API calls.

\section{Broader impacts}
\label{app:broader-impacts}

\methodsc{} reallocates LLM inference compute at test time; it does not train new models or change what tasks an LLM can perform, only how candidates are generated and selected for code-generation problems. The most plausible \emph{positive} impact is making weaker, cheaper code models more accurate for budget- or latency-constrained users (educational settings, low-resource deployments), which lowers the cost barrier to capable AI-assisted coding. The most plausible \emph{negative} impact is amplifying existing risks of LLM-generated code: confidently-wrong solutions selected by the consensus-based selector, propagation of insecure or biased code patterns from training data, and over-reliance on automated suggestions in domains where human review is essential. Two structural features of \methodsc{} bear directly on these risks: (i) execution-based selection filters out crashes but not silent semantic errors, so users should still treat selected candidates as suggestions rather than verified solutions; and (ii) the diversity-of-strategies design surfaces alternative algorithms, which can help users notice when the model's modal solution is wrong but does not guarantee any candidate is correct. We did not introduce new generative capabilities, so the harm surface of \methodsc{} is bounded by the base models it wraps; standard mitigations (sandboxed execution, human review, security audits of generated code) apply unchanged.

\end{document}